\documentclass{article}

\usepackage[preprint,nonatbib]{neurips_2022}

\usepackage[utf8]{inputenc} 
\usepackage[T1]{fontenc}    
\usepackage{hyperref}       
\usepackage{url}            
\usepackage{booktabs}       
\usepackage{amsfonts}       
\usepackage{amsmath}
\usepackage{bbm}
\usepackage{nicefrac}       
\usepackage{microtype}      
\usepackage{xcolor}         

\usepackage{graphicx}
\usepackage{multirow}

\usepackage{array}
\newcolumntype{C}[1]{>{\centering\let\newline\\\arraybackslash\hspace{0pt}}m{#1}}

\title{Consistent Video Instance Segmentation with Inter-Frame Recurrent Attention}

\author{%
  Quanzeng You\quad Jiang Wang \quad Peng Chu \quad Andre Abrantes \quad Zicheng Liu \\
  Microsoft Cloud \& AI\\
  \texttt{\{qyou, jiangwang, pengchu, abrantes, zliu\}@microsoft.com} \\
}

\begin{document}

\maketitle

\begin{abstract}
Video instance segmentation aims at predicting object segmentation masks for each frame, as well as associating the instances across multiple frames. 
Recent end-to-end video instance segmentation methods are capable of performing object segmentation and instance association together in a direct parallel sequence decoding/prediction framework. 
Although these methods generally predict higher quality object segmentation masks, they can fail to associate instances in challenging cases because they do not explicitly model the temporal instance consistency for adjacent frames.
We propose a consistent end-to-end video instance segmentation framework with Inter-Frame Recurrent Attention to model both the temporal instance consistency for adjacent frames and the global temporal context.
Our extensive experiments demonstrate that the Inter-Frame Recurrent Attention significantly improves temporal instance consistency while maintaining the quality of the object segmentation masks.
Our model achieves state-of-the-art accuracy on both YouTubeVIS-2019 (62.1\%) and YouTubeVIS-2021 (54.7\%) datasets.
In addition, quantitative and qualitative results show that the proposed methods predict more temporally consistent instance segmentation masks.


\end{abstract}
\begin{figure}[!hbtp]
    \centering
    \includegraphics[width=0.9\textwidth]{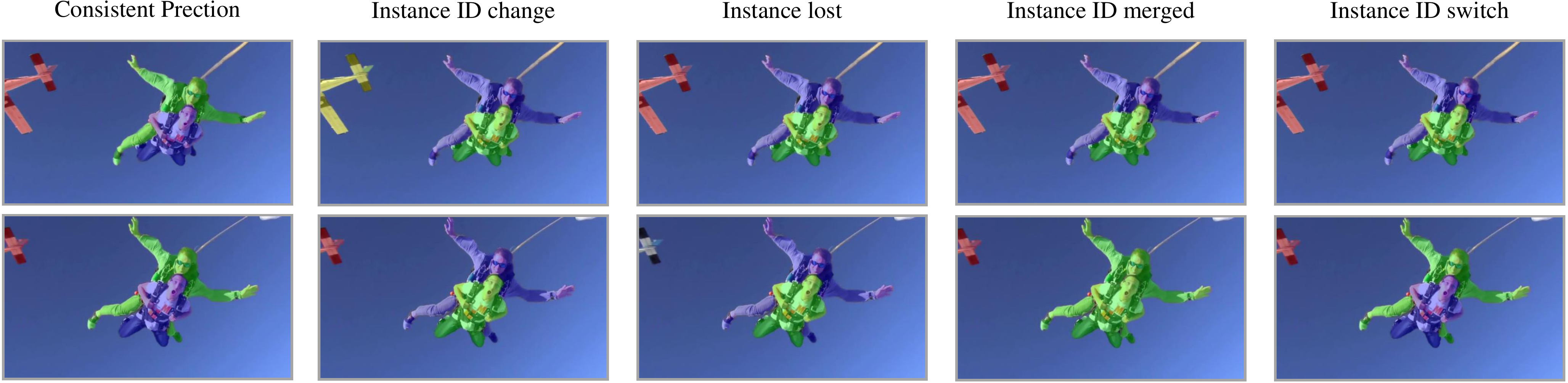}
    \caption{Examples of temporally consistent and inconsistent VIS results. Different mask colors represent different instance identities. The left-most column is the consistent result. Remaining columns are common inconsistent cases.}
    \label{fig:consistency}
    \vspace{-10pt}
\end{figure}

\section{Introduction}
Video instance segmentation (VIS)~\cite{yang2019video} is an emerging task in computer vision. It aims at simultaneously segmenting and tracking object instances in videos. 
Compared to image instance segmentation methods~\cite{he2017mask,bolya2019yolact,liu2018path}, VIS is more challenging because it does not only require predicting accurate segmentation results for each frame, but also needs to consistently associate the object instances across multiple frames. 


Previous VIS methods either independently segment each frame and associate per-frame instances~\cite{yang2019video,bertasius2020classifying,lin2020video,cao2020sipmask,yang2021crossover}, or predict the clip-level instance masks by taking the whole clip as input~\cite{athar2020stem,wang2021end} in an end-to-end manner.
Per-frame methods are usually more efficient, while end-to-end per-clip methods generally achieve better accuracy because they can predict the instance segmentation masks based on information from the whole video clip.
Recent methods like VisTR~\cite{wang2021end}, IFC~\cite{hwang2021video}, and Mask2Former~\cite{cheng2021mask2former,cheng2021mask2formervideo} demonstrate superior accuracy of end-to-end per-clip methods.

These end-to-end per-clip methods adopt a transformer-based end-to-end instance segmentation framework, where one object query is responsible for predicting the object category and segmentation masks of one instance in a video clip. 
The object category is predicted from the object query through a fully connected network. 
The segmentation masks of an instance are predicted by convolving the frame features independently with a dynamic kernel that is generated from the object query using a fully connected network and shared across all the frames. 

However, we observe that the end-to-end per-clip methods generally cannot maintain temporal instance consistency under challenging conditions, such as overlapping object instances or multiple object instances with the same category (see \figurename~\ref{fig:consistency} for examples), because they do not model the relationship of the instance segmentation masks in adjacent frames explicitly.
\figurename~\ref{fig:consistency} demonstrates common temporal inconsistencies in VIS.
The left-most column shows the consistent ground-truth identities and masks for three instances.  
The remaining columns demonstrate typical inconsistent VIS results, including identity change, instance lost, instance merge and instance identities switch.   
In videos, the adjacent frames are highly correlated, because the appearance and location of an object instance usually remain consistent in adjacent frames.
Incorporating this information explicitly is critical to achieve temporal instance consistency.
In this work, we propose an Inter-Frame Recurrent (IFR) attention framework to effectively model the correlation of adjacent frames, together with object appearance and clip-level global context.


In our IFR framework, we have two types of queries. \textit{Clip-level} queries predict object categories and summarize clip-level instance information.
\textit{Frame-level} queries predict the instance segmentation masks for the corresponding frames, and they model the specific information about each frame.
Several layers of IFR attention modules iteratively update clip-level and frame-level queries to model the temporal relationship.
An IFR attention module takes the frame-level queries of the previous frame, the clip-level queries, and the features of the current frame as inputs, and outputs the frame-level queries of the current frame through cross-attention and self-attention modules. 
The IFR attention module is capable of modeling both the relationship of the object instances in adjacent frames, as well as the relationship of the frame-level and clip-level object instances information. 
The clip-level queries aggregate the knowledge from all the frames by attending to all frame-level queries.
Then, the IFR attention modules in the next layer will utilize these clip-level queries as the clip-level query inputs.
We also design an auxiliary loss where the frame-level queries predict the masks of the same instance in other frames to further encourage temporal information propagation.


The proposed framework is simple yet effective as it models the appearance of the instances for each frame, the temporal correlation of the instances, and the clip-level global information. 
Compared with recent end-to-end approaches that only use global queries~\cite{cheng2021mask2former} for instance segmentation, our design leads to significantly more consistent temporal instance segmentation mask predictions, because it explicitly models the correlation of instances in adjacent frames through the IFR attention module. 


By using this framework, our approach sets the new state of the arts on both YouTubeVIS-2019 and YouTubeVIS-2021 datasets. 
In addition, we show, both quantitatively and qualitatively, that the proposed method generates more consistent instance segmentation masks while being more robust across multiple training runs.
We also propose a test-time augmentation (TTA) method to further boost the accuracy and robustness of the proposed method. 
We believe that the significantly improved temporal instance consistency of the proposed framework is not only beneficial to the overall instance segmentation accuracy, but also critical to the perceived quality of video instance segmentation.


\section{Related work}
\label{sec:related}
\paragraph{Image instance segmentation} Image instance segmentation attempts to classify an object's semantic category and recognize its pixels for a given image. 
One popular image instance segmentation method is a two-stage method called Mask R-CNN~\cite{he2017mask}, which detects each instance first and then recognizes the belonging pixels. 
Recently, due to their simplicity and competitive performance, one stage approaches~\cite{wang2020solo,tian2020conditional} are gaining popularity. 
In particular, the introduction of dynamic kernels~\cite{tian2020conditional} significantly boosted the performance of the one-stage approach~\cite{wang2020solov2}. 
More recently, the employment of Transformers~\cite{vaswani2017attention} and bipartite matching based ground truth assignments led to the success of end-to-end object detection~\cite{carion2020end,zhu2020deformable,liu2021dab}. 
In end-to-end object detection, each object query predicts both the category label and the bounding box of an object. 
This mechanism has been applied to end-to-end instance segmentation~\cite{cheng2021maskformer,hu2021istr,cheng2021mask2former} as well. 
In these approaches, object queries predict the semantic labels of the instances and produce dynamic kernels for mask prediction. 
Together with more powerful transformer-based backbones~\cite{liu2021swin}, these approaches set the new state of the art on image instance segmentation. 

\paragraph{Video instance segmentation} Video instance segmentation was introduced in~\cite{yang2019video} and involves segmenting instances at each frame while also tracking them across frames. 
Early studies~\cite{yang2019video,bertasius2020classifying,lin2020video,cao2020sipmask,yang2021crossover} independently locate instances in each frame and attempt to associate the instances across frames with tracking methods.
These per-frame based approaches are also known as tracking-by-detection. 
Later, STEm-Seg~\cite{athar2020stem} learns spatio-temporal embeddings from video clips, which are clustered and linked across frames to produce video-level instances. 
This approach is end-to-end trained, without an explicit tracking mechanism. 
Meanwhile, VisTR~\cite{wang2021end} builds their framework upon Transformers~\cite{vaswani2017attention}. 
Inspired by the end-to-end image instance segmentation, the instance segmentation results of each frame are produced by frame-level queries. 
Then, similarity learning is employed to group frame-level queries for instance tracking. 
STEm-Seg and VisTR also employ a clip matching mechanism for inference in long videos. 
To avoid the clip matching and computational complexity, Propose-Reduce~\cite{lin2021video} chooses to generate instance segmentation results on selected keyframes, and propagate the results to all the other frames. 
Each non-key frame receives and reduces the results from all keyframes to produce its result. 
This strategy balances performance and speed. 
The potential of end-to-end approaches is further explored in the recent state-of-the-art models~\cite{hwang2021video,wu2021seqformer,cheng2021mask2former}. 
Inter-Frame Communication (IFC)~\cite{hwang2021video} employs compact memory tokens to efficiently fuse all the frame tokens obtained from the transformer encoder. 
The output of the encoder is then fed to the decoder for video instance segmentation. 
SeqFormer~\cite{wu2021seqformer} employs frame-level box queries on each frame to generate frame-level mask queries for segmentation. 
The learning contains both mask prediction and object detection tasks. 
Mask2Former~\cite{cheng2021mask2former} directly uses all spatio-temporal tokens as inputs to the decoder to generate global mask queries, achieving a new state of the art.

\section{Approach}
\begin{figure}[!htbp]
	\centering
	\includegraphics[width=0.95\textwidth]{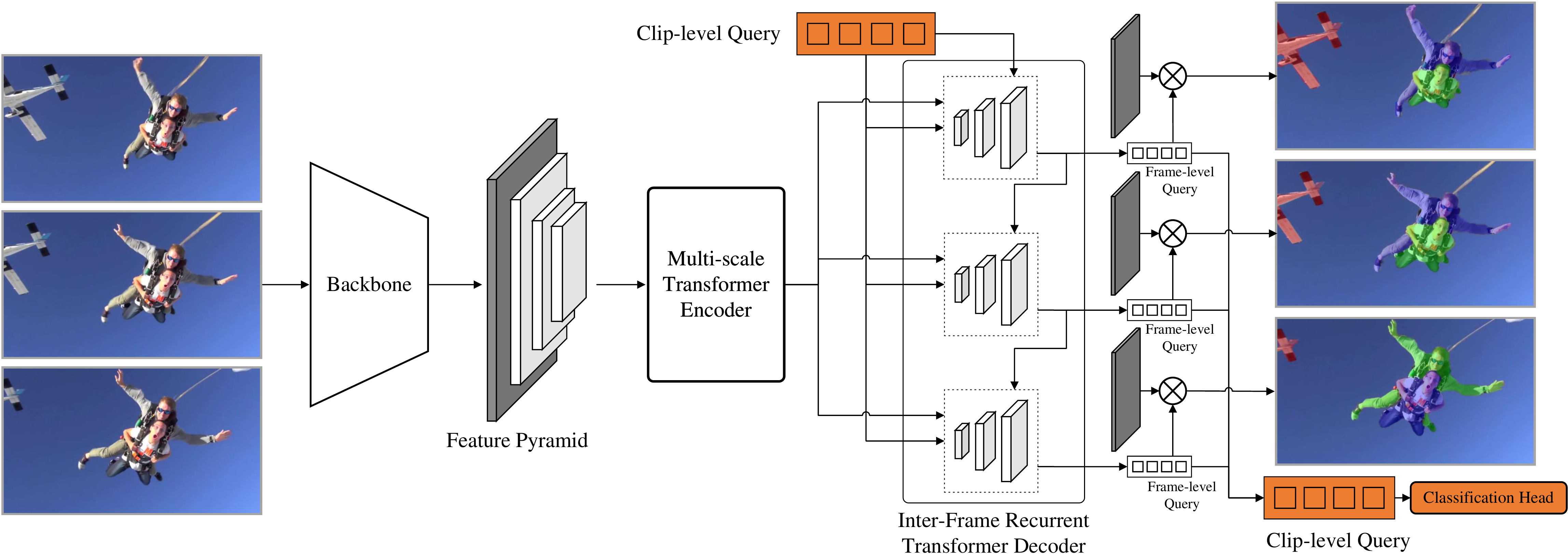}
	\caption{Illustration of our framework. Following recent studies, we also adopt multi-scale encoder to fuse multi-scale features. For each frame, we learn frame-level queries for mask prediction. The video-level queries, which aggregate the information across different frames, are utilized for the instance classification task.}
	\label{fig:framework}
\end{figure}

Our approach is based on the end-to-end framework that was initially proposed for object detection~\cite{carion2020end}.
Video instance segmentation involves two essential tasks: instance segmentation of each frame and instance association across frames. 
End-to-end approaches are able to achieve superior accuracy, because they allow both tasks to use all the information of the whole video by modeling the two essential tasks together. However, we found that state-of-the-art video instance segmentation methods exhibit temporally inconsistent instance segmentation masks because the instance association task is not well modeled. In this paper, we propose an Inter-Frame Recurrent (IFR) attention framework to significantly improve the modeling of instance association across frames in Transformer-based end-to-end video instance segmentation methods.
Similar to~\cite{cheng2021mask2former}, we employ a multi-scale Deformable encoder to fuse the pyramid of features from the backbone. 
The fused pyramid of features is used in IFR attention module to generate frame-level queries for mask prediction and clip-level queries for instance category classification. The IFR attention module simultaneously models the object appearance in each frame, the relationship of adjacent frames, and the clip-level context. 
Thus, it more effectively models instance association tasks by encouraging temporally consistent instance segmentation masks.
\figurename~\ref{fig:framework} shows the main components of the framework.

\subsection{Model framework}

\paragraph{Backbone} 
The backbone generates frame-level appearance features for each video frame.
Following the recent end-to-end work on video instance segmentation, we adopt image-based backbones for feature extraction. 
For an image $I$ with size of ${H \times W}$, the backbone produces a pyramid of feature maps with sizes of $H/4 \times W /4$, $H/8 \times W/8$, $H / 16 \times W/16$ and $H /32 \times W/32$. 

\paragraph{Multi-scale transformer encoder} 
Multi-scale transformer encoder fuses the information across the pyramid of feature maps in end-to-end instance segmentation tasks~\cite{hu2021istr,wu2021seqformer,cheng2021mask2former}. 
In this work, we adopt the memory-efficient multi-scale Deformable Transformer~\cite{zhu2020deformable} to fuse the pyramid of features, similar to ~\cite{cheng2021mask2former}.
The outputs of the multi-scale Deformable Transformer is a fused pyramid of features of the same size as the input.
Each scale of the fused pyramid of features is input to the instance segmentation decoder in a layer-by-layer fashion~\cite{cheng2021mask2former,cheng2021mask2formervideo}. 

\begin{figure}
	\centering
	\includegraphics[width=0.9\textwidth]{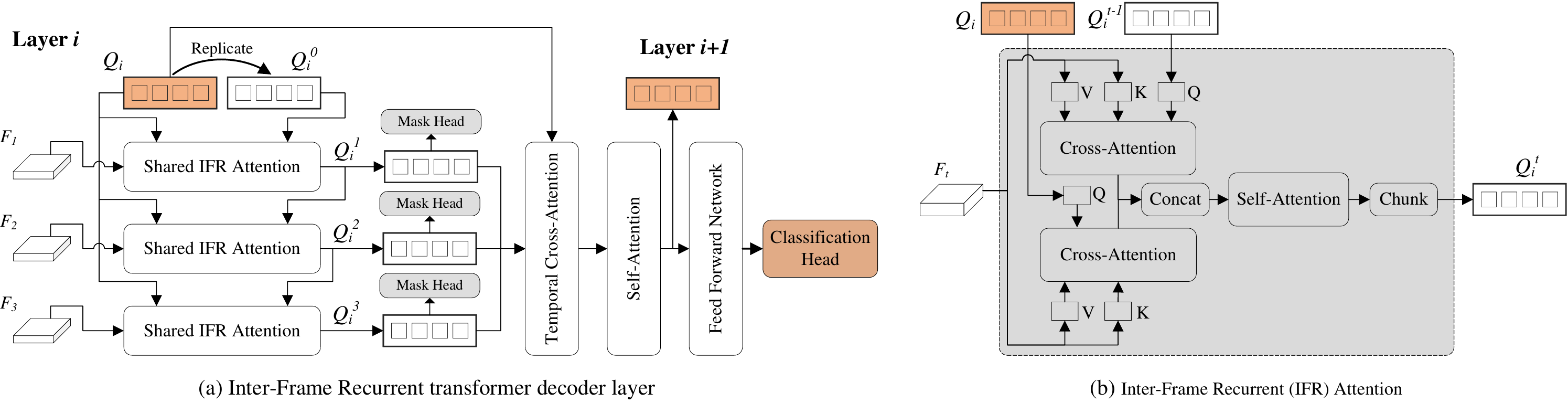}
	\caption{Architecture of our Inter-Frame Recurrent transformer decoder (for simplicity, we only show a clip of three frames). In the beginning stage of the $i$-th layer, we employ Inter-Frame Recurrent (IFR) attention to construct the \textit{frame-level} queries to predict the instance mask at each frame. The frame-level queries are aggregated with the temporal cross-attention module to generate the clip-level queries for video instance classification.
	}
	\label{fig:fat_attn}
\end{figure}

\paragraph{Inter-frame recurrent transformer decoder} 
For end-to-end transformer-based video instance segmentation, object queries play a critical role in modeling both instance appearance and temporal instance relationship. 
We propose the inter-frame recurrent transformer decoder as shown in~\figurename~\ref{fig:fat_attn}. 


For the $i$-th decoder layer, the set of \textit{clip-level} queries are denoted as $Q_i$. 
Each  clip also comes with $n$ sequential frame-level feature maps ($F_1, F_{2}, \dots, F_{n}$).  
The \textit{frame-level} queries for the $t$-th frame are denoted by $Q_i^{t}$.
Inter-frame recurrent attention module employs the \textit{clip-level} queries $Q_i$, the feature map $F_t$  , and the previous frame's \textit{frame-level} queries $Q_i^{t-1}$  to compute \textit{frame-level} queries $Q_i^{t}$ for mask prediction.  
The multi-head attention module, with hidden dimension of $d$, is employed 
\begin{equation}
	\text{Attn}(Q, K, V) = \text{Softmax}\left(\frac{(QK^T) \odot M}{\sqrt{d}}\right)V,
\end{equation}
where $M$ is an optional binary attention mask tensor and $\odot$ represents the element-wise multiplication between two tensors. 

In particular, given current feature map $F_{t}$, cross-attention module is utilized to compute the updated frame-level and clip-level queries from $Q_i$ and $Q_i^{t-1}$. 
Then, both groups of queries are concatenated to form $Q_i^{t'}$:
\begin{equation}
	Q_i^{t'} = \left[ \text{Attn}(Q_i^{t-1}, F_{T+t}W^K, F_{T+t}W^V), \text{Attn}(Q_i, F_{T+t}W^K, F_{T+t}W^V)\right].\label{eqn:local}
\end{equation}
$Q_i^{t'}$ goes through a self-attention module. 
The eventual queries $Q_i^t$ are obtained by splitting $Q_i^{t'}$ and keeping the first group of queries.
\begin{equation}
	Q_i^t = \text{Chunk}(\text{Attn}(Q_i^{t'}, Q_i^{t'}{W'}^K, Q_i^{t'}{W'}^V)).
\end{equation}
In this way, the frame-level queries $Q_i^t$ do not only utilize the appearance features from $F_{t}$ and clip-level features, 
but also leverage the local temporal information from $Q_i^{t-1}$. 
The local temporal information is crucial for encouraging temporally consistent instance segmentation mask predictions.
At the start of a clip $t=1$, we initialize the previous frame-level query $Q_i^0$ to be a copy of clip-level query $Q_i$ (as shown in \figurename~\ref{fig:fat_attn}). 

The above steps compute the \textit{frame-level} queries for each frame. 
Next, we update \textit{clip-level} queries $Q_i$ by aggregating the knowledge in all the frame-level queries $\mathbf{Q} = \{Q_i^{1}, \dots Q_i^{n}\}$. 
This is achieved by using temporal cross-attention, where $Q_i$s are the query, and all frame-level queries $\mathbf{Q}$ are used as key and value. 
Another self-attention module on the output of the temporal cross-attention generates the clip-level queries $Q_{i+1}$. 
$Q_{i+1}$ is used for the instance classification task, as well as the clip-level queries for the $(i+1)$-th layer.

\subsection{Instance segmentation training loss}
Similar to previous work, one ground-truth video instance is associated with one object query with Hungarian algorithm.
The cost of assigning $i$-th ground-truth instance to $\pi_i$-th clip-level query is 
\begin{equation}
	\mathcal{C}_\text{match}=-\lambda_c \hat{p}_{\pi_i}(c_i) + \lambda_m \mathcal{C}_m(m_i, \hat{m}_{\pi_i})  + \lambda_d ( 1 - \textsc{Dice}(m_i, \hat{m}_{\pi_i})), \label{eqn:matching}
\end{equation}
where $\hat{p}_{\pi_i}$ is the predicted classification probability of clip-level query $\pi_i$, $\mathcal{C}_m(\cdot, \cdot)$ computes the binary cross-entropy between predicted and ground-truth masks, 
and the \textsc{Dice} is the dice coefficient defined in~\cite{milletari2016v}. The optimal assignment $\pi$ is computed as minimizing the cost for all $M$ ground-truth instances in a training clip.

\paragraph{Training loss} The training loss is similar to the above matching cost. 
It consists of the classification loss $\mathcal{L_{\text{cls}}}$, the cross-entropy mask loss $\mathcal{L}_{\text{mask}}$ and the \textsc{Dice} loss $\mathcal{L}_{\textsc{Dice}}$.  

As discussed above, Inter-Frame Recurrent attention module is employed to generate \textit{frame-level} queries for mask prediction. 
Although the information from the other frames in the clip can be propagated through \textit{clip-level} queries and previous frame's  \textit{frame-level} queries, we find it is beneficial to add an auxiliary cross-frame mask prediction loss to further utilize temporal information, where each \textit{frame-level} queries are used to predict the masks of the same instance in the other frames in the same training clip.
Our training loss can be represented as:
\begin{equation}
	\mathcal{L}  = \sum_{i = 1}^{N} \lambda_c\mathcal{L}_\text{cls}(c_i, \hat{p}_{\pi_i}) + \mathbbm{1}_{c_{\pi_i}\ne \oslash} \left[\lambda_m  \bar{\mathcal{L}}_m(m_i, \hat{m}_{\pi_i}) + \lambda_{D} \bar{\mathcal{L}}_D (m_i, \hat{m}_{\pi_i}) \right],\label{eqn:loss} 
\end{equation}
where $\lambda_c$, $\lambda_m$ and $\lambda_D$ are the loss weights, $\mathbbm{1}_{c_{\pi_i}\ne \oslash}$ controls only queries that matched with ground-truth will have mask and \textsc{Dice} loss. 
For the classification loss $\mathcal{L}_\text{cls}$, we employ Focal loss~\cite{lin2017focal}, which demonstrates performance gain for instance segmentation as well~\cite{zhu2020deformable}. 
Both $\bar{\mathcal{L}}_m$ and $\bar{\mathcal{L}}_D$ consider the cross-frame mask predictions as well. 
More specifically, we have 
\begin{align}
	\bar{\mathcal{L}}_m (m_i, \hat{m}_{\pi_i}) & =	\sum_{t_1} \left[\mathcal{L}_m(m_i^{t_1}, \hat{m}^{t_1}_{\pi_i}) + \lambda_{e}\sum_{t_2, t_2 \ne t_1}\mathcal{L}_m(m^{t_2}_i, \hat{m}_{\pi_i}^{t_1}) \right] \label{eqn:loss_mask}\\
	\bar{\mathcal{L}}_D (m_i, \hat{m}_{\pi_i}) & = \sum_{t_1} \left[( 1 - \textsc{Dice}(m_i^{t_1}, \hat{m}_{\pi_i}^{t_1})) + \lambda_{e}\sum_{t_2, t_2 \ne t_1}( 1 - \textsc{Dice}(m^{t_2}_i, \hat{m}_{\pi_i}^{t_1})) \right]\label{eqn:loss_dice}, 
\end{align}
where $\lambda_e$ is the weight for cross-frame mask loss, $m_i^t$ and $\hat{m}_{\pi_i}^t$ are the ground-truth mask of the $i$-th instance and the matched predicted mask at frame $t$ of the given training clip. 

\paragraph{Test time augmentation (TTA)} Test time augmentation is widely used in computer vision tasks~\cite{krizhevsky2012imagenet,he2016deep}. 
TTA feeds multiple augmented testing images or videos to the network to produce multiple predictions for a testing image or video.
The final result is obtained by aggregating all the predicted results using a post-processing method, such as Non-Maximum Suppression (NMS)~\cite{neubeck2006efficient} in object detection.
For instance segmentation, TTA is rarely applied due to the difficulty in merging multiple results. 
We propose a TTA mechanism for instance segmentation by applying the matching mechanism in the training stage (Eq.(~\ref{eqn:matching})).
If we have $n$ different instance segmentation results $\{R_1, R_2, \dots, R_n\}$ from the same video, we randomly choose one result $R_p$ as the pivot. 
All other results are matched to $R_p$ based on the matching cost in Eq.(~\ref{eqn:matching}) so that each instance in $R_p$ is matched to $n-1$ instances.
The aggregated result for an instance contains the averaged classification probabilities and instance mask probabilities for the corresponding $n$ matched instances.
A similar strategy was adopted for clip-level instance tracking in~\cite{hwang2021video}.

\section{Experiments}
\label{sec:exp}
Following previous studies~\cite{yang2019video}, we evaluate the proposed approach on YouTubeVIS-2019 and YouTubeVIS-2021 datasets (released with Creative Commons Attribution 4.0 License).\footnote{\url{https://youtube-vos.org/dataset/vis/}}
The two datasets contain similar videos, annotations, and evaluation metrics.
YouTubeVIS-2021 contains more videos, improved categories, and roughly twice as much as video instance annotations compared to the 2019 version.
We adopt the mean of average precision (AP) over all classes are the main metric, which is similar to the AP metric in image instance segmentation.
We also report AP50 and AP75, where IoU thresholds of $0.5$ and $0.75$ are used when calculating AP respectively, as well as average recall (AR).
All the metrics are generated using the official evaluation servers of YouTubeVIS-2019  and YouTubeVIS-2021 datasets.

\begin{table}[!htb]
  \caption{Evaluation results on \textbf{YouTubeVIS-2019} validation split. }
  \label{tab:ytvis2019}
  \centering
  \small
  \begin{tabular}{l|l|l|lcc|cc}
    \toprule
    Method & Backbone & Data & AP & AP50 & AP75 &AR1 & AR10\\
    \midrule
    \multirow{2}{*}{SeqFormer} & ResNet-50 & Y + COCO & 47.4 & 69.8 & 51.8 & 45.5 & 54.8\\
    & ResNet-101 & Y + COCO & 49.0 & 71.1 & 55.7 & 46.8 & 56.9 \\
    \midrule
    \multirow{2}{*}{Mask2Former} & ResNet-50 & Y & 46.4$\pm$0.8 & 68.0 & 50.0 & - & -\\
     & ResNet-101 &  Y & 49.2$\pm$0.7 & 72.8 & 54.2  & - & -\\
    \midrule
    \multirow{2}{*}{Ours} & ResNet-50 & Y & 46.5$\pm$0.8 & 68.4 & 50.1 & 46.3 & 55.9\\
    & ResNet-101 & Y & 49.3$\pm$0.4 & 75.2 & 59.7 & 48.5 & 58.2 \\
    \midrule
    \midrule
    SeqFormer* & Swin-L & Y + COCO & 59.4 & 82.1 & 66.4 & 51.7 & 64.4 \\
    \midrule
    \multirow{4}{*}{Mask2Former} & Swin-T & Y & 51.5$\pm$0.7 & 75.0 & 56.6 & - & -\\
    & Swin-S &  Y & 54.3$\pm$0.7 & 79.0 & 58.8 & - & -\\
    & Swin-B &  Y & 59.5$\pm$0.7 & 84.3 & 67.2 & - & -\\
    & Swin-L-200 &  Y & 60.4$\pm$0.5 & 84.4 & 67.0 & - & -\\
    \midrule 
    \multirow{6}{*}{Ours} & Swin-T &  Y & 52.8$\pm$0.3 & 75.2 & 59.7 & 50.3 & 61.7\\
    & Swin-S &  Y & 56.2$\pm$0.1 & 79.9 & 62.3 & 52.5 & 63.8\\
    & Swin-B &  Y & 61.3$\pm$0.4 & 84.6 & 68.4 & 54.3 & 66.7\\
    & Swin-L &  Y & 62.1$\pm$0.2 & 84.9 & 69.2 & 55.1 & 67.1\\
    & Swin-L-200 & Y &  61.2$\pm$0.3 & 84.6 & 69.3 & 54.8 & 67.3\\
    \cmidrule{2-8} 
    & Swin-L (TTA) & Y& 62.6$\pm$0.1 & 84.8 & 69.3 & 55.5 & 68\\
    \bottomrule
  \end{tabular}
\flushleft\scriptsize{\quad *For Swin-L, input resolution is 360 for shorter size, where 480 is used for Mask2Former and our model.}\\
\end{table}

\subsection{Training setup}
\label{sec:setup}
We implement our model based on IFC~\cite{hwang2021video} with Apache 2.0 license and Mask2Former~\cite{cheng2021mask2former} with Creative Commons Attribution-NonCommercial 4.0 International License, using Detectron2 framework~\cite{wu2019detectron2} under Apache License 2.0. 
All models are trained with a node of 8 NVIDIA V100 GPUs. 
Following the common recipe in~\cite{hwang2021video,cheng2021mask2formervideo}, we initialize our model with weights pre-trained on COCO dataset because both YouTubeVIS2019 and YouTubeVIS2021 datasets are relatively small-scale.
Since pretraining the inter-frame recurrent attention modules is not possible with an image instance segmentation dataset, we randomly initialize these modules.
During training, we randomly sample clips~\cite{hwang2021video,cheng2021mask2formervideo} from each video.
The number of frames for each video is set to 2, and the shorter side of the training clip is resized to $360$ or $480$, following ~\cite{cheng2021mask2former}.
The learning rate for the backbone is $1/10$ of the learning rate of other parts of the network.
We follow the same setting in~\cite{cheng2021mask2former}: $\lambda_c = 2$, $\lambda_m = 5$ and $\lambda_{D} = 5$ in Eq.(\ref{eqn:loss}).
$\lambda_e = 0.3$ in Eq.(\ref{eqn:loss_mask}).
During inference, the model takes the whole video and predicts the video instance segmentation for all the frames.

\subsection{Main results}
In this section, we report the results on both YouTubeVIS 2019 and YouTubeVIS 2021 datasets. 
All metrics are obtained by training our models on the training splits independently five times and averaging the evaluation metrics.
We mainly compare with SeqFormer~\cite{wu2021seqformer} and Mask2Former~\cite{cheng2021mask2former,cheng2021mask2formervideo}, which are state-of-the-art Transformer-based end-to-end approaches.

\paragraph{YouTubeVIS 2019 results}  \tablename~\ref{tab:ytvis2019} summarizes the results on the validation split of YouTubeVIS 2019 dataset. 
For CNN-based backbone, our results are comparable with SeqFormer~\cite{wu2021seqformer} and Mask2Former~\cite{cheng2021mask2former}. 
Our approach demonstrated large performance gains using Swin Transformer~\cite{liu2021swin} backbones.
Our experiments also show that the proposed model is able to achieve better performance using fewer queries (100 queries compared to 200 queries in Mask2Former~\cite{cheng2021mask2former}). 
In particular, our model with Swin-L backbone and 100 queries achieve $62.1\pm0.2$ AP, which is almost 2 percent better than the previous state-of-the-art end-to-end method.

In test time augmentation, we resize the testing videos to four different input resolutions during inference, 
and average the matched predictions. 
The ``Swin-L (TTA)'' is the result of using TTA on Swin-L backbone.
The TTA strategy improves AP and reduces the variances of the results.

\paragraph{YouTubeVIS 2021 results} YouTubeVIS 2021 is a more challenging updated version of the YouTubeVIS 2019 dataset. 
The results are summarized in \tablename~\ref{tab:ytvis2021}. 
Our models demonstrate the best results across all different backbones.
Compared with Mask2Former, our model achieves a much smaller variance with higher \textit{AP}. 
This dataset is more challenging, thus TTA shows a larger gain compared with the results on YouTube VIS 2019 dataset. 

\begin{table}[!htbp]
  \caption{Evaluation results on \textbf{YouTubeVIS-2021} validation split. }
  \label{tab:ytvis2021}
  \small
  \centering
  \begin{tabular}{l|l|l|lcc|cc}
    \toprule
    Method & Backbone & Data & AP & AP50 & AP75 & AR1 & AR10 \\
    \midrule
    SeqFormer~\cite{wu2021seqformer} & ResNet-50 & Y + COCO & 40.5 & 62.4 & 43.7  & 36.1 & 48.1 \\
    \midrule
    \multirow{2}{*}{Mask2Former} & ResNet-50 & Y & 40.6$\pm$0.7 & 60.9 & 41.8 & - & -\\
     & ResNet-101 & Y & 42.4$\pm$0.6 & 65.9 & 45.8  & - & -\\
    \midrule
    \multirow{2}{*}{Ours} & ResNet-50 & Y & 41.8$\pm$0.2 & 63.0 & 44.9 & 39.2 & 48.8\\
    & ResNet-101 & Y & 42.9$\pm$0.1 & 64.7 & 46.2 & 40.0 & 50.5\\
    \midrule
    \midrule
    SeqFormer & Swin-L & Y+COCO & 51.8 & 74.6 & 58.2 & 42.8 & 58.1  \\
    \midrule
    \multirow{4}{*}{Mask2Former} & Swin-T & Y & 45.9$\pm$0.6 & 68.7 & 50.7 & - & -\\
    & Swin-S & Y & 48.6$\pm$0.4 & 77.2 & 52.0 & - & -\\
    & Swin-B & Y & 52.0$\pm$0.6 & 76.5 & 54.2 & - & -\\
    & Swin-L & Y & 52.6$\pm$0.7 & 76.4 & 57.2 & - & -\\
    \midrule 
    \multirow{7}{*}{Ours} & Swin-T & Y & 47.4$\pm$0.9 & 69.2 & 52.5 & 41.7 & 54.3\\
    & Swin-S & Y & 49.4$\pm$0.2 & 71.2 & 54.0 & 43.5 & 55.9\\
    & Swin-B & Y & 53.9$\pm$0.2 & 76.4 & 60.2 & 45.1 & 58.8\\
    & Swin-L & Y & 54.7$\pm$0.2 & 76.8 & 61.2 & 44.4 & 58.6\\
    & Swin-L (200) & Y & 54.0$\pm$0.1 & 75.8 & 60.1 & 45.2 & 58.8\\
    \cmidrule{2-8}
    & Swin-L (TTA) & Y & 55.6$\pm$0.05 & 77.3 & 61.8 & 45.7 & 59.6\\
    \bottomrule
  \end{tabular}
\end{table}

\subsection{Ablation study for cross-frame mask prediction loss}
\begin{figure}[!htbp]
	\centering
	\includegraphics[width=0.95\textwidth]{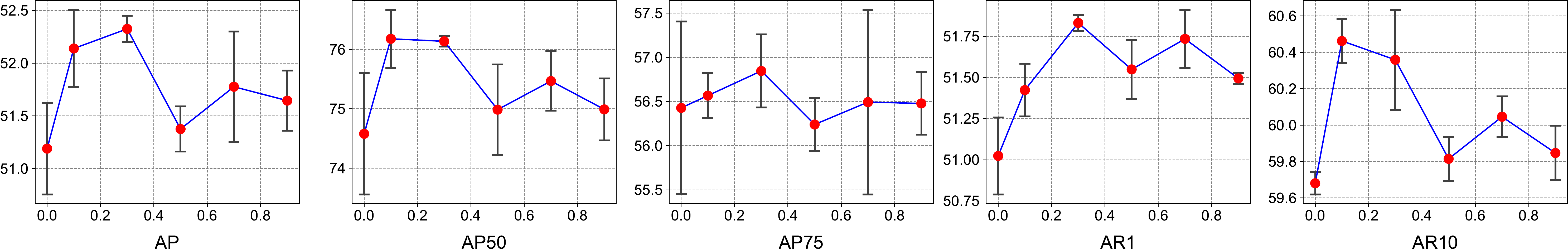}
	\caption{Results of our propose model with different loss weights $\lambda_e$ (Eq.(\ref{eqn:loss_mask}) and Eq.(\ref{eqn:loss_dice})). All models are trained with ResNet-50 as the backbone. For each $\lambda_e$, we run five experiments. The mean and variance (indicated by the error bar) of each parameter setting for different metrics are visualized.}
	\label{fig:cmask}
\end{figure}

We perform ablation study for the auxiliary cross-frame mask prediction loss (Eq.(\ref{eqn:loss_mask}) and Eq.(\ref{eqn:loss_dice})). 
To study the impact of this loss, we randomly select $400$ videos from the YouTubeVIS2019 training set as the ablation study validation set, 
and train the proposed model with ResNet-50 backbone using different values of $\lambda_e$ on the remaining training data. 
For each $\lambda_e$, we run five independent training. The means and variances of the evaluation metrics are visualized in \figurename~\ref{fig:cmask}. 
The results suggest that the proposed consistency mask loss can improve the performance of AP compared with the results without using the loss ($\lambda_e = 0$), although a too large $\lambda_e$ leads to worse results.
Overall, the best performance is achieved when $\lambda_e = 0.3$. 
We choose $\lambda_e = 0.3$ in all of our experiments.

\subsection{Evaluation of temporal consistency}
In this section, we design an experiment to evaluate the temporal consistency of our model through tracking metrics.

We calculate a tight bounding box that encloses each instance mask as the bounding box for each instance object in both the predicted and ground-truth results, 
and employ the widely used tracking metrics in multiple object tracking (MOT) task~\cite{bernardin2008evaluating,ristani2016performance}. 
The tracking metrics can characterize the instance-level temporal inconsistency, such as false-positive and false-negative instance tracks. 
Trajectory ID-related metrics include ID precision (IDP), ID recall (IDR), IDF1, and identity switches between tracklets. 
Multiple object tracking accuracy (MOTA), which measures the overall accuracy of tracker and detection, is another important metric. 
We employ the open-sourced implementation\footnote{\url{https://github.com/cheind/py-motmetrics} released under MIT License.} with an IoU threshold of $0.5$, which controls the computation of the metrics and is commonly adopted in tracking tasks. 
We conduct this experiment on the same ablation study data splitting as in \figurename~\ref{fig:cmask}, because we need to access the ground-truth annotations.

\begin{table}[!htbp]
	\caption{Comparison between different approaches in terms of the tracking metrics, which measure the performance of associating instances across frames. We compare the two approaches on both YouTubeVIS (YTVIS) 2019 and YouTubeVIS (YTVIS) 2021 datasets.}
	\label{tab:tracking}
	\small
	\centering
	\begin{tabular}{l|l|l|ccccc}
		\toprule
		Method & Backbone & Dataset & IDF1$\uparrow$ &MOTA$\uparrow$ & IDP$\uparrow$ & IDR$\uparrow$ & IDs$\downarrow$\\
		\midrule
		\multirow{2}{*}{Mask2Former} & ResNet-50 & YTVIS19& 62.4\% & 14.6 & 54.7\% & 74.3\% & 202 \\
		& Swin-L (200) & YTVIS19& 66.7\% & 23.0 & 57.1\% & 80.3\% & 218 \\
		\midrule
		\multirow{3}{*}{Ours} & ResNet-50 &YTVIS19& 72.2\% & 47.6 & 74.7\% & 70.0\%  & 50\\
		& Swin-L & YTVIS19&77.8\% & 57.5 & 78.2\% & 77.4\% & 48 \\
		& Swin-L (200) & YTVIS19&78.3\% & 58.7 & 79.2\% & 77.4\% & 26 \\
		\midrule
		\midrule
		\multirow{2}{*}{Mask2Former} & ResNet-50 & YTVIS21 & 55.1\% & -1.6 & 47.3\% & 66.2\% & 790 \\
		& Swin-L (200) & YTVIS21 & 61.7\% & 14.5 & 53.0\% & 73.9\% & 729 \\
		\midrule
		\multirow{3}{*}{Ours} & ResNet-50 & YTVIS21 & 64.1\% & 36.4 & 70.1\% & 59.1\% & 141 \\
		& Swin-L & YTVIS21 & 70.8\% & 47.5 & 74.5\% & 67.4\% & 123 \\
		& Swin-L (200) & YTVIS21 & 71.1\% & 48.7 & 76.4\% & 66.4\%  & 94 \\
		\bottomrule
	\end{tabular}
\end{table}
\tablename~\ref{tab:tracking} shows the results of Mask2Former and ours with ResNet-50 and Swin-L backbones on both datasets. 
Our models demonstrate significant improvements in terms of the tracking metrics. 
The scores on IDF1, IDs and MOTA suggest the effectiveness of our proposed model in terms of preserving consistent identity ID for instances across all the video frames. 
\subsection{Quantitative results visualization}
\begin{figure}
	\centering
	\includegraphics[width=0.9\textwidth]{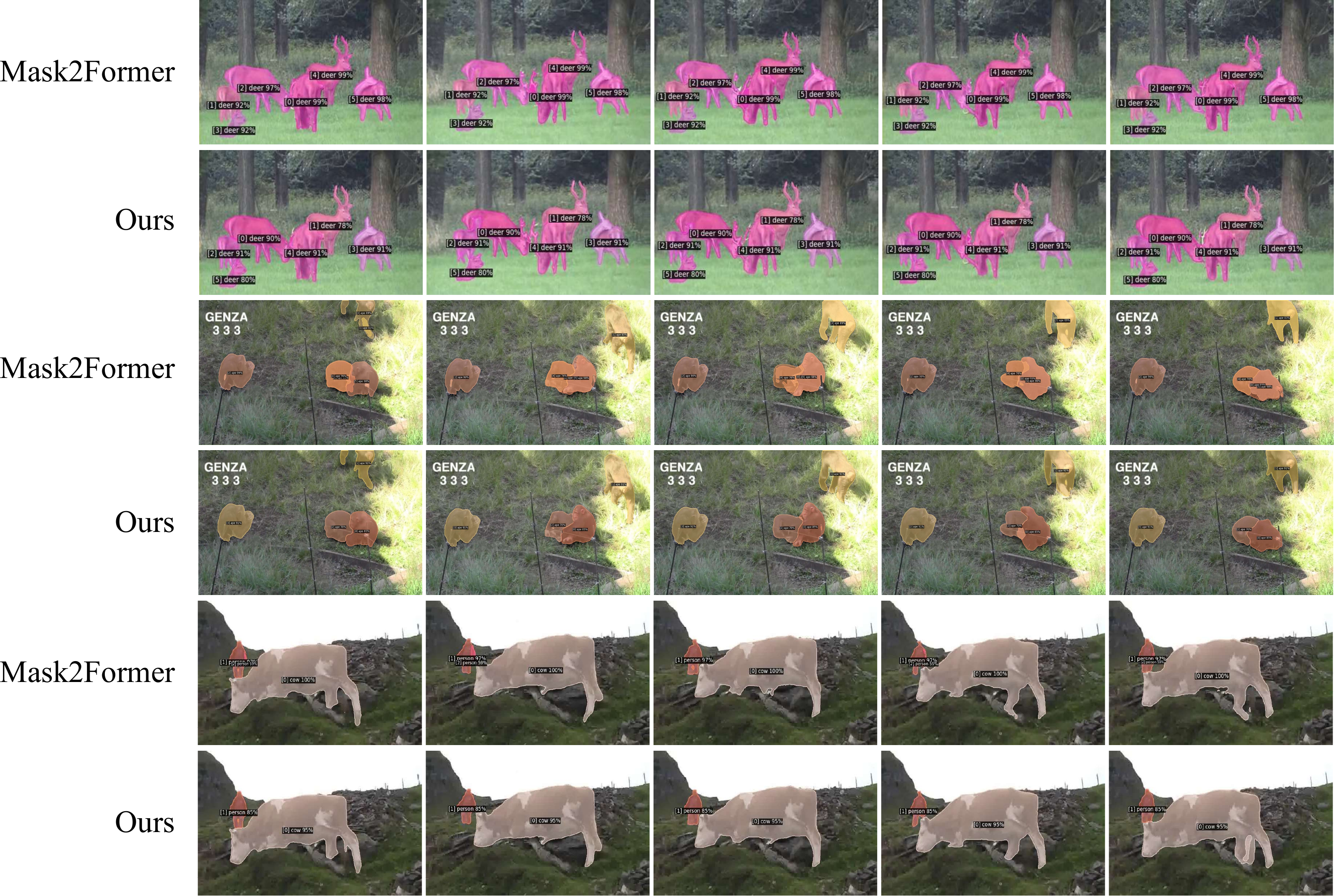}
	\caption{Visualization results of our model and Mask2Former on selected multi-instance sequences. All results are from the model with Swin-L backbone.}
	\label{fig:vis}
\end{figure}
To visually compare the prediction results of the proposed method and Mask2Former, we select several instance segmentation results, as shown in~\figurename~\ref{fig:vis}, from both models trained with Swin-L backbone. 
For the first sequence, both approaches demonstrate consistent video instance prediction results. 
All instances are accurately segmented within each frame and correctly tracked across frames. 
For this video, it is easier for the model to predict temporally consistent instances, because all the ``deers'' stay almost at the same poses with only slight movements. 
For the second clip, there are multiple ``apes'': one sits still, two of them play with each other and the last one walks by.
As suggested by the results, Mask2Former generates duplicate predictions for the walking ape and the two playing apes. The instance IDs for these apes also switched. 
In comparison, our results assign unique and consistent instance ID for each ape.
The last sequence contains two instances with different categories: person and cow. The cow passes by the person. 
The person is assigned as two different instances by Mask2Former, while our results allocate one instance for the person. 
In practice, we found that temporal inconsistencies significantly degrade the perceived video segmentation quality, because humans can easily observe these temporal inconsistencies. 
These results show that our approach can assign consistent instance identities across frames.

\subsection{Limitation}
\label{sec:limit}
Our models are too computationally expensive to be deployed for real-time inference.
We can implement per-clip inference, but the speed still cannot meet the requirements of real-time applications.
\section{Conclusion}
\label{sec:conc}
Video instance segmentation is a relatively new vision task. 
It involves instance segmentation within each frame and instance association across frames. 
In this work, we attempt to improve the instance association quality by simultaneously employing Inter-Frame Recurrent (IFR) attention for integrating frame-specific features and propagating video-level temporal features. 
In addition, we also propose to include the cross-frame mask prediction loss for discriminative frame-level query learning. 
In this way, the proposed approach achieves the new state-of-the-art performance on both YouTubeVIS 2019 and YouTubeVIS 2021 datasets. 
Inspired by the matching strategy in associating ground truth with each query, we also propose the test time augmentation strategy, which further benefits the inference results. 
Our work fuses the temporal information for video instance segmentation primarily in the head of the network. 
In the future, we hope temporal information can be more effectively aggregated in network backbones for dense prediction video tasks. 
Our models are trained on video data, which may contain social biases and the training process can cause environmental costs. 
We will examine and alleviate these social impacts in the future.

{
\small

\bibliographystyle{plain} 
\bibliography{refs} 

\begin{thebibliography}{10}

\bibitem{athar2020stem}
Ali Athar, Sabarinath Mahadevan, Aljosa Osep, Laura Leal-Taix{\'e}, and Bastian
  Leibe.
\newblock Stem-seg: Spatio-temporal embeddings for instance segmentation in
  videos.
\newblock In {\em European Conference on Computer Vision}, pages 158--177.
  Springer, 2020.

\bibitem{bernardin2008evaluating}
Keni Bernardin and Rainer Stiefelhagen.
\newblock Evaluating multiple object tracking performance: the clear mot
  metrics.
\newblock {\em EURASIP Journal on Image and Video Processing}, 2008:1--10,
  2008.

\bibitem{bertasius2020classifying}
Gedas Bertasius and Lorenzo Torresani.
\newblock Classifying, segmenting, and tracking object instances in video with
  mask propagation.
\newblock In {\em Proceedings of the IEEE/CVF Conference on Computer Vision and
  Pattern Recognition}, pages 9739--9748, 2020.

\bibitem{bolya2019yolact}
Daniel Bolya, Chong Zhou, Fanyi Xiao, and Yong~Jae Lee.
\newblock Yolact: Real-time instance segmentation.
\newblock In {\em Proceedings of the IEEE/CVF international conference on
  computer vision}, pages 9157--9166, 2019.

\bibitem{cao2020sipmask}
Jiale Cao, Rao~Muhammad Anwer, Hisham Cholakkal, Fahad~Shahbaz Khan, Yanwei
  Pang, and Ling Shao.
\newblock Sipmask: Spatial information preservation for fast image and video
  instance segmentation.
\newblock In {\em European Conference on Computer Vision}, pages 1--18.
  Springer, 2020.

\bibitem{carion2020end}
Nicolas Carion, Francisco Massa, Gabriel Synnaeve, Nicolas Usunier, Alexander
  Kirillov, and Sergey Zagoruyko.
\newblock End-to-end object detection with transformers.
\newblock In {\em European conference on computer vision}, pages 213--229.
  Springer, 2020.

\bibitem{cheng2021mask2formervideo}
Bowen Cheng, Anwesa Choudhuri, Ishan Misra, Alexander Kirillov, Rohit Girdhar,
  and Alexander~G Schwing.
\newblock Mask2former for video instance segmentation.
\newblock {\em arXiv preprint arXiv:2112.10764}, 2021.

\bibitem{cheng2021mask2former}
Bowen Cheng, Ishan Misra, Alexander~G. Schwing, Alexander Kirillov, and Rohit
  Girdhar.
\newblock Masked-attention mask transformer for universal image segmentation.
\newblock 2022.

\bibitem{cheng2021maskformer}
Bowen Cheng, Alexander~G. Schwing, and Alexander Kirillov.
\newblock Per-pixel classification is not all you need for semantic
  segmentation.
\newblock In {\em NeurIPS}, 2021.

\bibitem{he2017mask}
Kaiming He, Georgia Gkioxari, Piotr Doll{\'a}r, and Ross Girshick.
\newblock Mask r-cnn.
\newblock In {\em Proceedings of the IEEE international conference on computer
  vision}, pages 2961--2969, 2017.

\bibitem{he2016deep}
Kaiming He, Xiangyu Zhang, Shaoqing Ren, and Jian Sun.
\newblock Deep residual learning for image recognition.
\newblock In {\em Proceedings of the IEEE conference on computer vision and
  pattern recognition}, pages 770--778, 2016.

\bibitem{hu2021istr}
Jie Hu, Liujuan Cao, Yao Lu, ShengChuan Zhang, Yan Wang, Ke~Li, Feiyue Huang,
  Ling Shao, and Rongrong Ji.
\newblock Istr: End-to-end instance segmentation with transformers.
\newblock {\em arXiv preprint arXiv:2105.00637}, 2021.

\bibitem{hwang2021video}
Sukjun Hwang, Miran Heo, Seoung~Wug Oh, and Seon~Joo Kim.
\newblock Video instance segmentation using inter-frame communication
  transformers.
\newblock {\em Advances in Neural Information Processing Systems}, 34, 2021.

\bibitem{krizhevsky2012imagenet}
Alex Krizhevsky, Ilya Sutskever, and Geoffrey~E Hinton.
\newblock Imagenet classification with deep convolutional neural networks.
\newblock {\em Advances in neural information processing systems}, 25, 2012.

\bibitem{lin2020video}
Chung-Ching Lin, Ying Hung, Rogerio Feris, and Linglin He.
\newblock Video instance segmentation tracking with a modified vae
  architecture.
\newblock In {\em Proceedings of the IEEE/CVF Conference on Computer Vision and
  Pattern Recognition}, pages 13147--13157, 2020.

\bibitem{lin2021video}
Huaijia Lin, Ruizheng Wu, Shu Liu, Jiangbo Lu, and Jiaya Jia.
\newblock Video instance segmentation with a propose-reduce paradigm.
\newblock In {\em Proceedings of the IEEE/CVF International Conference on
  Computer Vision}, pages 1739--1748, 2021.

\bibitem{lin2017focal}
Tsung-Yi Lin, Priya Goyal, Ross Girshick, Kaiming He, and Piotr Doll{\'a}r.
\newblock Focal loss for dense object detection.
\newblock In {\em Proceedings of the IEEE international conference on computer
  vision}, pages 2980--2988, 2017.

\bibitem{liu2021dab}
Shilong Liu, Feng Li, Hao Zhang, Xiao Yang, Xianbiao Qi, Hang Su, Jun Zhu, and
  Lei Zhang.
\newblock Dab-detr: Dynamic anchor boxes are better queries for detr.
\newblock In {\em International Conference on Learning Representations}, 2021.

\bibitem{liu2018path}
Shu Liu, Lu~Qi, Haifang Qin, Jianping Shi, and Jiaya Jia.
\newblock Path aggregation network for instance segmentation.
\newblock In {\em Proceedings of the IEEE conference on computer vision and
  pattern recognition}, pages 8759--8768, 2018.

\bibitem{liu2021swin}
Ze~Liu, Yutong Lin, Yue Cao, Han Hu, Yixuan Wei, Zheng Zhang, Stephen Lin, and
  Baining Guo.
\newblock Swin transformer: Hierarchical vision transformer using shifted
  windows.
\newblock In {\em Proceedings of the IEEE/CVF International Conference on
  Computer Vision}, pages 10012--10022, 2021.

\bibitem{milletari2016v}
Fausto Milletari, Nassir Navab, and Seyed-Ahmad Ahmadi.
\newblock V-net: Fully convolutional neural networks for volumetric medical
  image segmentation.
\newblock In {\em 2016 fourth international conference on 3D vision (3DV)},
  pages 565--571. IEEE, 2016.

\bibitem{neubeck2006efficient}
Alexander Neubeck and Luc Van~Gool.
\newblock Efficient non-maximum suppression.
\newblock In {\em 18th International Conference on Pattern Recognition
  (ICPR'06)}, volume~3, pages 850--855. IEEE, 2006.

\bibitem{ristani2016performance}
Ergys Ristani, Francesco Solera, Roger Zou, Rita Cucchiara, and Carlo Tomasi.
\newblock Performance measures and a data set for multi-target, multi-camera
  tracking.
\newblock In {\em European conference on computer vision}, pages 17--35.
  Springer, 2016.

\bibitem{tian2020conditional}
Zhi Tian, Chunhua Shen, and Hao Chen.
\newblock Conditional convolutions for instance segmentation.
\newblock In {\em European Conference on Computer Vision}, pages 282--298.
  Springer, 2020.

\bibitem{vaswani2017attention}
Ashish Vaswani, Noam Shazeer, Niki Parmar, Jakob Uszkoreit, Llion Jones,
  Aidan~N Gomez, {\L}ukasz Kaiser, and Illia Polosukhin.
\newblock Attention is all you need.
\newblock {\em Advances in neural information processing systems}, 30, 2017.

\bibitem{wang2020solo}
Xinlong Wang, Tao Kong, Chunhua Shen, Yuning Jiang, and Lei Li.
\newblock Solo: Segmenting objects by locations.
\newblock In {\em European Conference on Computer Vision}, pages 649--665.
  Springer, 2020.

\bibitem{wang2020solov2}
Xinlong Wang, Rufeng Zhang, Tao Kong, Lei Li, and Chunhua Shen.
\newblock Solov2: Dynamic and fast instance segmentation.
\newblock {\em Advances in Neural information processing systems},
  33:17721--17732, 2020.

\bibitem{wang2021end}
Yuqing Wang, Zhaoliang Xu, Xinlong Wang, Chunhua Shen, Baoshan Cheng, Hao Shen,
  and Huaxia Xia.
\newblock End-to-end video instance segmentation with transformers.
\newblock In {\em Proceedings of the IEEE/CVF Conference on Computer Vision and
  Pattern Recognition}, pages 8741--8750, 2021.

\bibitem{wu2021seqformer}
Junfeng Wu, Yi~Jiang, Wenqing Zhang, Xiang Bai, and Song Bai.
\newblock Seqformer: a frustratingly simple model for video instance
  segmentation.
\newblock {\em arXiv preprint arXiv:2112.08275}, 2021.

\bibitem{wu2019detectron2}
Yuxin Wu, Alexander Kirillov, Francisco Massa, Wan-Yen Lo, and Ross Girshick.
\newblock Detectron2.
\newblock \url{https://github.com/facebookresearch/detectron2}, 2019.

\bibitem{yang2019video}
Linjie Yang, Yuchen Fan, and Ning Xu.
\newblock Video instance segmentation.
\newblock In {\em Proceedings of the IEEE/CVF International Conference on
  Computer Vision}, pages 5188--5197, 2019.

\bibitem{yang2021crossover}
Shusheng Yang, Yuxin Fang, Xinggang Wang, Yu~Li, Chen Fang, Ying Shan, Bin
  Feng, and Wenyu Liu.
\newblock Crossover learning for fast online video instance segmentation.
\newblock In {\em Proceedings of the IEEE/CVF International Conference on
  Computer Vision}, pages 8043--8052, 2021.

\bibitem{zhu2020deformable}
Xizhou Zhu, Weijie Su, Lewei Lu, Bin Li, Xiaogang Wang, and Jifeng Dai.
\newblock Deformable detr: Deformable transformers for end-to-end object
  detection.
\newblock In {\em International Conference on Learning Representations}, 2020.

\end{thebibliography}
}

\appendix

\section{Appendix}
\subsection{Per-clip inference}
\begin{table}[!htbp]
	\caption{Evaluation results of using per-clip inference on both datasets of using Swin-L backbone. The clip length $T = v$ indicates per-video inference.}
	\label{tab:clip}
	\small
	\centering
	\begin{tabular}{l|l|c|lcc|cc}
		\toprule
		Dataset & Backbone & Clip length (T) & AP & AP50 & AP75 &AR1 & AR10\\
		\midrule
		\multirow{4}{*}{YTVIS2019} & Swin-L & v & 62.1$\pm$0.2 & 84.9 & 69.2 & 55.1 & 67.1\\
		& Swin-L (TTA) & v & 62.6$\pm$0.1 & 84.8 & 69.3 & 55.5 & 68\\
		& Swin-L-CLIP & 2 & 61.6$\pm$0.6 & 84.3 & 68.0 & 54.3 & 66.0 \\
		& Swin-L-CLIP & 5 & 61.8$\pm$0.2 & 85.5 & 68.0 & 54.3 & 65.9 \\
		\midrule
		\multirow{4}{*}{YTVIS2021} & Swin-L & v & 54.7$\pm$0.2 & 76.8 & 61.2 & 44.4 & 58.6\\
		& Swin-L (TTA) & v & 55.6$\pm$0.05 & 77.3 & 61.8 & 45.7 & 59.6\\
		& Swin-L-CLIP & 2 & 56.6$\pm$0.4 & 79.2 & 63.0 & 45.6 & 60.3 \\
		& Swin-L-CLIP & 5 & 56.3$\pm$0.3 & 78.4 & 62.5 & 45.6 & 59.7 \\
		\bottomrule
	\end{tabular}
\end{table}
We also implement the clip-tracking strategy~\cite{hwang2021video} to conduct per-clip inference. 
We experiment with different clip-length (T) settings on both datasets.
\tablename~\ref{tab:clip} compares the results. 
All per-clip results are also computed by averaging multiple runs. 
On YTVIS2019 dataset, per-clip inference leads to slightly worse results. 
However, on the YTVIS2021 dataset, per-clip inference exhibits better results. 
We conjecture that better results are potentially related to the almost doubled number of instances in YTVIS 2021 version dataset. 
With longer clip length and more instances, the inference gets more challenging when associating across instances. 
This will be further investigated and mitigated in the future. 

However, per-clip inference increases the variances of the results. 
Especially, with $T=2$, the variance is doubled on YTVIS2021 dataset and tripled on YTVIS2019 dataset. 

\end{document}